# FACE IDENTIFICATION BY MEANS OF A NEURAL NET CLASSIFIER

Virginia Espinosa-Duró, Marcos Faúndez-Zanuy *
Departamento de Electrónica y Automática
* Departamento de Telecomunicaciones y Arquitectura de computadores
Escuela Universitaria Politécnica de Mataró, Adscrita a la UPC
Avda. Puig y Cadafalch 101-111 08303 Mataró (BCN), Spain
tel:34 (93) 757 44 04 fax:34 (93) 757 05 24
e-mail: espinosa@eupmt.es, faundez@eupmt.es

This paper describes a novel face identification method that combines the eigenfaces theory with the Neural Nets. We use the eigenfaces methodology in order to reduce the dimensionality of the input image, and a neural net classifier that performs the identification process.

The method presented recognizes faces in the presence of variations in facial expression, facial details and lighting conditions. A recognition rate of more than 87% has been achieved, while the classical method of Turk and Pentland achieves a 75.5%

## INTRODUCTION

In recent years the problem of face recognition has attracted considerable attention. Human faces represent one of the most common visual patterns in our environment. For a computer, a picture of a human face is just a map of pixels of different gray (or color) levels. Thus in order to recognize an individual using his face it has to represent a human face in an intelligent way, that is, to represent a face image as a feature vector of reasonably low dimension and high discriminating power. Developing this representation is the main challenge for automated face-recognition researchers and developers. Great progress has been made toward developing computer vision algorithms that can recognize individuals based on their facial images The existing approaches for face recognition may be classified into two categories: *holistic* and *analytic*. The holistic approaches consider the global properties of the pattern, while the second one considers a set of geometrical features of the face. Exist two main divisions of these second one approaches, based on feature vectors extracted from profile silhouettes, and from a frontal view of the face.

This paper presents an holistic approach for face recognition .The face recognition system that we propose consists of two main stages:

To use the eigenfaces method like a feature extractor obtaining an output result represented as a pattern vector, whose components are computed one at a time to allow the quickest possible response, and a neural net classifier that performs the identification process.

Our system is compared against the classical method that consists on computing an error measure between the parameterized input face (the input image projected into the face subspace) and the parameterizations of all the faces modeled in the training process (that form the models of each people).

All the experiments described here, have been executed on the faces provided by the ORL face database. The next section of this paper describes this face database.

### Database

The database used is the ORL (Olivetti Research Laboratory) database of faces. This database contains a set of face images taken between April 1992 and April 1994 at ORL. The database was used in the context of a face recognition project carried out in collaboration with the Speech, Vision and Robotics Group of the Cambridge University Engineering Department.

There are ten different images of each of 40 distinct subjects. For some subjects, the images were taken at different times, varying the lighting, facial expressions (open/closed eyes, smiling / not smiling) and facial details (glasses / no glasses). All the images were taken against a dark homogeneous background with the subjects in an upright, frontal position (with tolerance for some side movement).

## THE EIGENFACE APROACH

Turk and Pentland [2,3], propose an eigenface system which projects face images onto a feature space that spans the significant variations among

________________________________________________________________________________

known face images using the Karhunen-Loéve Transform. It is an orthogonal lineal transform of the signal that concentrates the maximum information of the signal with the minimum number of parameters using the minimum square error (MSE). The significant features are known as eigenfaces, because they are the eigenvectors (principal components) of the set of images. The projection operation characterizes an individual face by a weighted sum of the eigenface features, and so to recognize a particular face it is only necessary to compare these weigths to those of known individuals.

### Computing Eigenfaces. Used Method

The computations assume a training sequence of M images $\{I_1, I_2, \ldots, I_M\}$ of size NxN (Although the images must not be squared we will assume that they are squared without lost of generality). This set of images belongs to a number smaller or equal to M persons (if there is one or more face images for each person).

The eigenfaces method [2] can be summarized in the following steps:

- The images are re-arranged into a one-dimensional vector of $1xN^2$ components.
- To compute the average face:

$$\Psi = \frac{1}{M}\sum_{n=1}^{M} I_n$$

- The average face is substracted to all the vectors:

$$\Phi_i = I_i - \Psi$$

- To compute the covariance matrix, with the following expression, where T means the transpose of the vector.

$$C = \frac{1}{M}\sum_{n=1}^{M}\Phi_n \Phi_n^T$$

- This matrix can be expressed as:

$$C = AA^T$$ where the matrix $A = [\Phi_1 \Phi_2 \ldots \Phi_M]$.

- This matrix is diagonalized and the u eigenvectors ($u \ll N^2$) associated to the u greatest eigenvalues are selected. Each of these eigenvectors is named eigenface. Thus, it is possible to obtain a more compacted representation. The geometric interpretation is that each face is approximated by a linear combination of eigenvectors of the selected subspace.

- The matrix C has a dimension $N^2 \times N^2$ elements. Thus, for a N=92x112 pixels/image the diagonalization of the matrix implies a high computational burden and memory requirements.

- For this reason, usually the computations are done with the matrix $A^T A$, which is smaller. In this case, using the eigenvectors $\vec{v}_l = [v_{l1}, v_{l2}, \ldots, v_{lM}]$ can be derived an approximated expression for the eigenvectors $\vec{u}_l = [u_{l1}, u_{l2}, \ldots, u_{lN^2}]$:

$$\vec{u}_l = \sum_{k=1}^{M}\vec{v}_{lk}\Phi_k \qquad l=1,\ldots,M$$

- For the testing process, the coordinates of the test face ($\Omega$) must be compared against the coordinates of the modeled persons ($\Omega_K$). This can be done projecting the test face into the subspace computed previously. It is not needed to use the M eigenvectors obtained. It is possible to use the M'<M more significatives. The projection of the input face (I) over the M' directions of the vectorial space can be implemented in the following way:

$$\omega_k = \vec{u}_k^T (I - \Psi)$$

- The coordinates $\omega$ form the vector $\Omega^T = [\omega_1, \omega_2, \ldots, \omega_{M'}]$ that parameterizes the input face.

### Face Recognition

Recognition is performed by finding the training face that minimizes the face distance with respect to the input test face. In other terms, the identification of the test image is done locating the database entry, whose weights are closest (in euclidean distance) to the weights of the face.

### ARTIFICIAL NEURAL NETWORK

We have used a feedforward Multi Layer Perceptron as a classifier instead of the classical Mean Square Error classifier. This is because The performance of the MSE is not enough good when the dimensionality of the vectors is high, which is precisely the case of face recognition.

Special effort is devoted to the neural net training, because it is the key factor to achieve a good performance. Main subjects are:
- Number of layers and neurons in each layer.
- Selection of training, testing, and validation databases.
- Number of epochs, training algorithm, parameter normalization, initialization of weights (multi-start), etc.

## Learning Algorithm

We have considered among the most effective approaches to machine learning when the data includes complex sensory input such as images:

a) *The Gradient Descent w/Momentum & Adaptive learning rate Backpropagation* training function that updates weight and bias values according to gradient descent momentum and an adaptive learning rate.

b) *The Levenberg-Marquardt training algorithm,* that computes the appoximate Hessian matrix (dim n by n). In our simulations we have seen that it is not possible to obtain this matrix, which is close to singular. This is because the number of parameters of the neural net is greater than the amount of information (training vectors). Thus, we have used the first method.

## Generalization

One of the problems that occurs during neural network training is called *overfitting*. The error on the training set is driven to a very small value, but when new data is presented to the network the error is large. The network has memorized the training examples, but it has not learned to generalize to new situations. Two solutions to the overfitting problem have been evaluated:

a) Regularization
The regularization involves modifying the performance function, which is normally chosen to be the sum of squares of the network errors on the training set. So, this technique helps take the mystery out of how to pick the number of neurons in a network and consistently leads to good networks that are not *overtrained*.

The classical Mean Square Error (MSE) implies the computation of:

$$MSE = \frac{1}{N} \sum_{i=1}^{N} (t_i - a_i)^2$$

where t,a are the N dimensional vectors of the test input and the model, respectively.
The regularization uses the following measure:

$$MSEREG = \gamma MSE + (1-\gamma) \frac{1}{n} \sum_{j=1}^{n} w_j^2$$

Thus, it includes one term proportional to the modulus of the weights of the neural net.

b) Early stopping with Validation
Early stopping is a technique based on dividing the data into three subsets. The first subset is the training set used for computing the gradient and updating the network weights and biases. The second one is the validation test. The error on the validation set is monitorized during the training process. The validation error will normally decrease during the initial phase of training, as does the training set error. However, When the network begins to overfit the data, the error of validation set will typically begin to rise. When the validation error increases for a specified number of iterations, the training is stopped, and the weigths and biases at the minimum of the validation error are returned.

We have observed that the validation does not suppose a considerable improvement in the recognition rates (using the test database like the validation test). Perhaps this is because of the reduced number of training vectors, so we have used the regularization method for our final results.

## RESULTS

In this section we will present an exhaustive study about the recognition rates as function of the involved parameters. Our results have been obtained with the ORL database in the following situation:40 persons, faces 1 to 5 for training, and faces 6 to 10 for testing.

Figure 1 shows the recognition rates as function of the number of projections, that is, how many parameters represent each image (the dimensionality of the vectors). The number of persons is 40 and the number of faces for each person used for training is 5, so the maximal number of coefficients is 40x5=200.

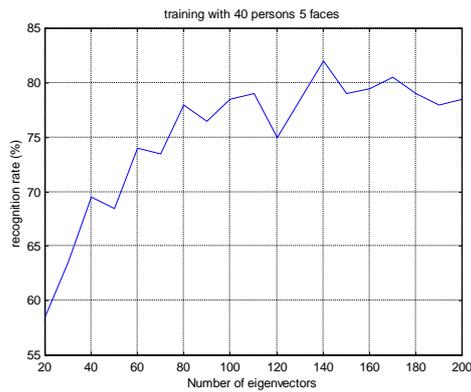

*Figure 1 Recognition rates as function of eigenvectors*

Based upon figure 1 we have selected a dimension of 80, because it offers a good compromise between dimensionality and recognition rates.

Table 1 summarizes the most relevant results:

| Method | Recognition rate |
|---|---|
| Eigenfaces (dim=80) | 75.5% |
| Eigenfaces (dim=200) | 78% |
| Neural net 80x30x40 | 84% |
| Neural net 80x40x40 | 87% |

Table 1: Recognition rates for several situations

### Computing the models

The training has been done in the following way:
1. Neural net: when the input is a genuine face, the output (target of the nnet) is fixed to 1. When the input is an impostor face, the output is fixed to –1.
2. Classical eigenface approach: we compute 5 different models for each person. Thus, in the test process the identity of the input face is guessed if it is assigned to one of his five face models.

In our simulations we have obtained that a suitable number of epochs is 4000.

Figures 2 and 3 show the histograms of the distances for genuine and impostor faces. There are two interesting commentaries: for the neural net, there is a preponderance of the negative responses. This is because of the most part of the training vectors are inhibitory. Thus, the nnet tends to learn that "all is inhibitory". Although this, it is possible to discriminate between classes. On the other hand, the histograms of the genuine faces for the classical eigenface approach have been represented in two situations. The first one shows the distance from the input faces to the five models of the same person. The last one shows the distance from the input faces to the nearest face of the same person.

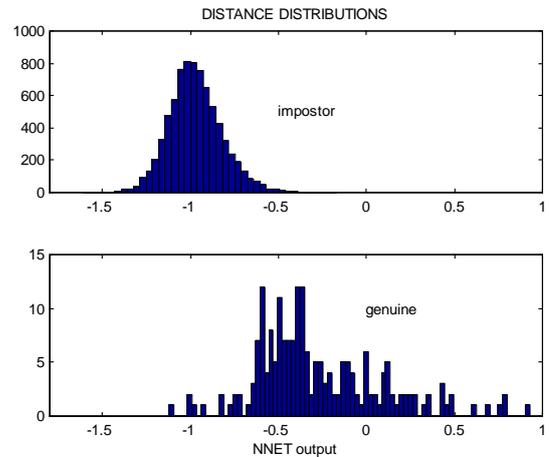

*Figure 2: Histograms outputs of the nnet*

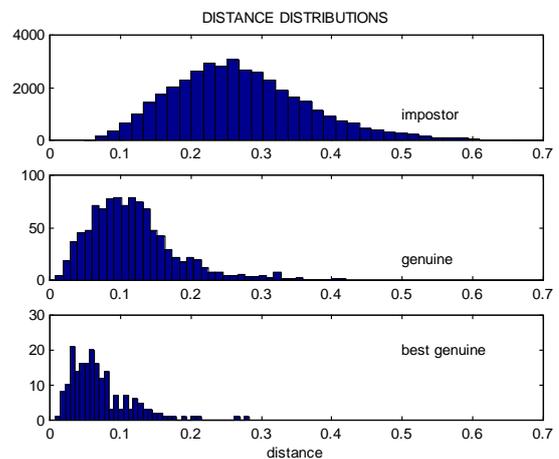

*Figure 3: Histograms distances of the classical method*

### CONCLUSIONS

The main conclusions are the following:
- We have proven that with a properly training phase, the accuracy of the neural net classifier outperforms the classical approach based on MSE.
- We believe that the neural net has a greater generalization capability for recognizing faces not used during training.
- The classical method is a particular case of classification based on a predefined error measure (typically Mean Square Error). On the other hand the neural net has more feasibility in order to adapt its transfer function to the optimal desired function.

___________________________________________________________________________________
____